\documentclass[wcp]{jmlr}
\jmlrvolume{1}
\jmlryear{2015}
\jmlrworkshop{ICML 2015 AutoML Workshop}

\usepackage{float}
\restylefloat{table}

\title[AutoCompete]{AutoCompete: A Framework for Machine Learning Competitions}

  \author{\Name{Abhishek Thakur} \Email{thakur@aisbi.de}\and
   \Name{Artus Krohn-Grimberghe} \Email{artus@aisbi.de}\\
   \addr AISBI, University of Paderborn, Germany}

\begin{document}

\maketitle

\begin{abstract}
In this paper, we propose AutoCompete, a highly automated machine learning framework for tackling machine learning competitions. This framework has been learned by us, validated and improved over a period of more than two years by participating in online machine learning competitions. It aims at minimizing human interference required to build a first useful predictive model and to assess the practical difficulty of a given machine learning challenge. The proposed system helps in identifying data types, choosing a machine learning model, tuning hyper-parameters, avoiding over-fitting and optimization for a provided evaluation metric. We also observe that the proposed system produces better (or comparable) results with less runtime as compared to other approaches.
  
\end{abstract}
\begin{keywords}
auto-machine learning, predictive modelling
\end{keywords}

\section{Introduction}
\label{sec:intro}


In the industry, business analysts are usually not concerned with the algorithms, feature selection, feature engineering or selection of appropriate hyperparameters. All they want is a fast track to a highly accurate predictive model which they can apply with minimum knowledge and effort on their problems and datasets. To satisfy this need, many "one-click" machine learning platforms have emerged that specifically target those users. Platforms such as Google Predict and BigML take the dataset as input from the end user and provide them with a predictive model for the dataset and a web service to consume it but that is beyond the scope of this paper.

In machine learning research, this topic has arrived under the umbrella term AutoML that subsumes and integrates disjunct areas of research such as identification of the problem (classification/regression, identifying the type of data, types of features and selection of features). Besides connecting these areas, AutoML also offers the possibility to delve into meta-learning where generalization from one dataset would tell which approaches can also be applied successfully on similar datasets. 

We propose a system that automates a lot of the classical machine learning cycle and tries to build a predictive model without (or with a very little) human interference. 

With the advent of various popular machine learning competition platforms such as CodaLab \citep{codalab}, Kaggle \citep{kaggle}, DrivenData \citep{dd}, etc., it is now easy to gather a broad set of distinct datasets with different features that represent real-world machine learning problems. Our hypothesis is that, an AutoCompete framework to ease the life of a business user should be able to benefit from the learnings of a human expert on a large enough set of ML competitions at least in the form of codified knowledge. 

\begin{figure}[H]
\centering
\includegraphics[scale = 0.29]{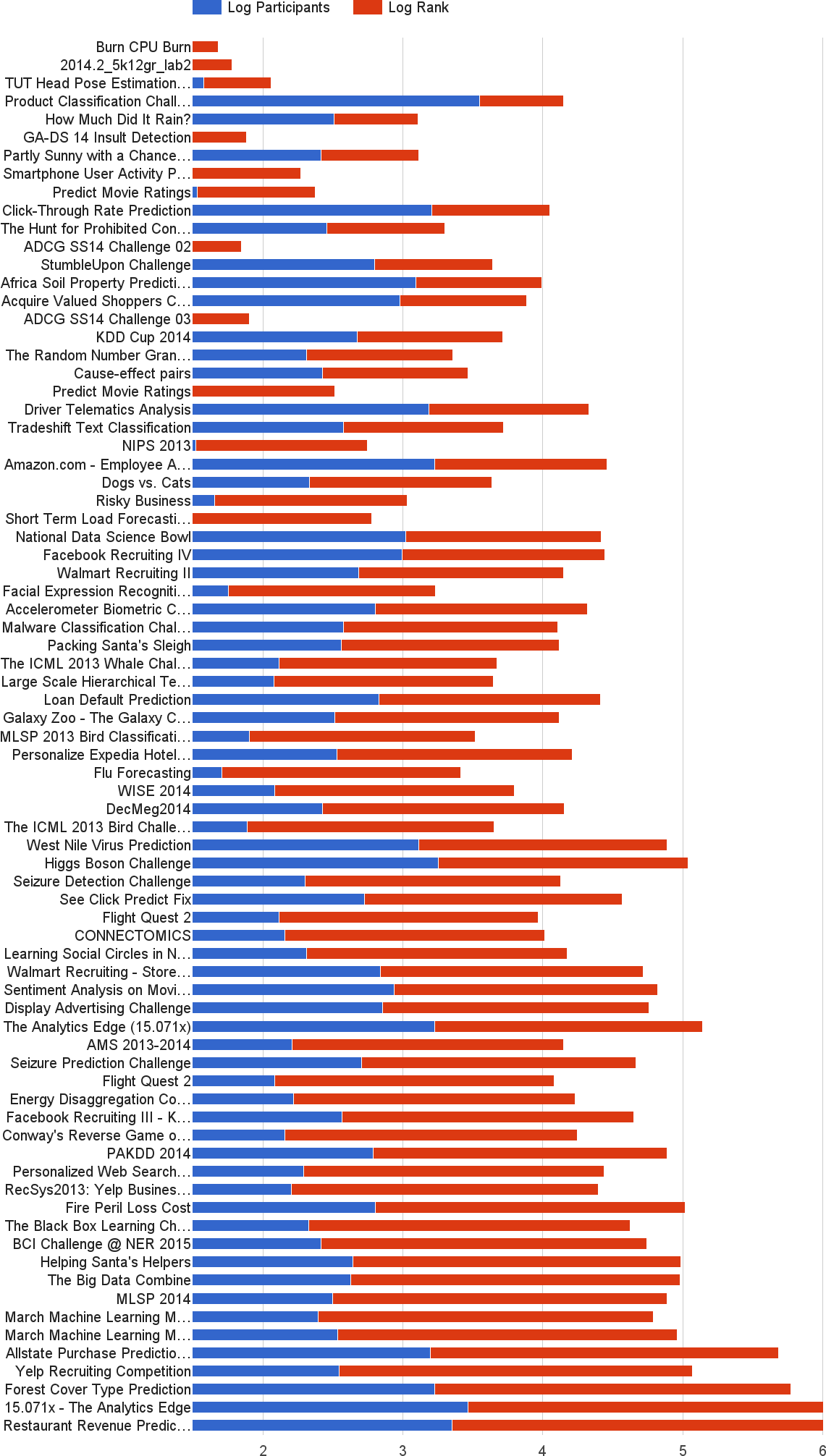}
\caption{Number of participants v/s rank obtained in various machine learning competitions (log scaled). Only the data for Kaggle is shown here.}
\end{figure}

We use this knowledge to train the AutoCompete system tackle different types of datasets. The system has been trained from knowledge acquired over a period of more than two years and more than 100 machine learning competitions. Figure 1 shows the performance of our human expert supported by earlier versions of this framework in selected machine learning competitions. It is to be noted that the good performance is obtained as a result of both the human expert and the AutoCompete framework. The framework was developed over time and new pipelines were added according to the requirement of the datasets seen by the human expert.

This paper is divided into five sections. Section 2 discusses our approach to AutoCompete. In section 3 we discuss the main components of the proposed AutoCompete framework followed by section 4 which discusses results on standard datasets and comparison with other such systems. Section 5 gives the conclusion and future work along with the feasibility of such a system.

\section{Base Framework}
\label{sec:baseframework}

As most competition datasets follow this layout, our current AutoCompete system works only with datasets in tabular format. Such a dataset can be defined as a set $X$ and a vector $y$, where, every row in $X$ represents a sample and every corresponding row in $y$ is the label (or output feature) of that sample. Every column of $X$ is an input feature. The proposed system is presently unable to deal with datasets in other formats. If such a dataset is encountered, a human expert is invited to convert the format which can then be used for predictive modelling using the proposed AutoCompete system.

The most important components of the proposed AutoCompete system, as depicted in Figure 2 are the ML Model Selector and Hyper-parameter Selector. In addition to these, there is a data splitter, data type identifier, feature stacker, decomposition tools and feature selector. 

Once a tabular data is fed into the AutoCompete system, the very first step taken by it is splitting the dataset into training and validation sets. If a classification task is encountered, the dataset is split in a stratified manner, such that both the training and validation set have the same distribution of labels. The validation set is always kept separate from any transformations being used on the training set and is not touched at any point in the pipeline.

All the transformations on the training set are saved and then applied on the validation set in the end. This ensures that the system is not over-fitting and the models thus produced as a result of the AutoCompete pipeline generalize on unseen datasets. Once the splitting is done, the type of features are identified. The data types for every feature can be supplied by the user. However, if the manually specified data types are not available, the system distinguishes between different features on its own by applying basic heuristics. For example, if text dataset is encountered, AutoCompete system will deploy natural language processing based algorithms and text transformers. For others, data type is identified and appropriate transformations are used. Each transformation is then fed through a feature selection mechanism which in turn sends the selected features and the transformation pipeline through model selector and hyper-parameter selector. The transformation and the model with the best performance is used in the end.

\begin{figure}[H]
\centering
\includegraphics[scale = 0.25]{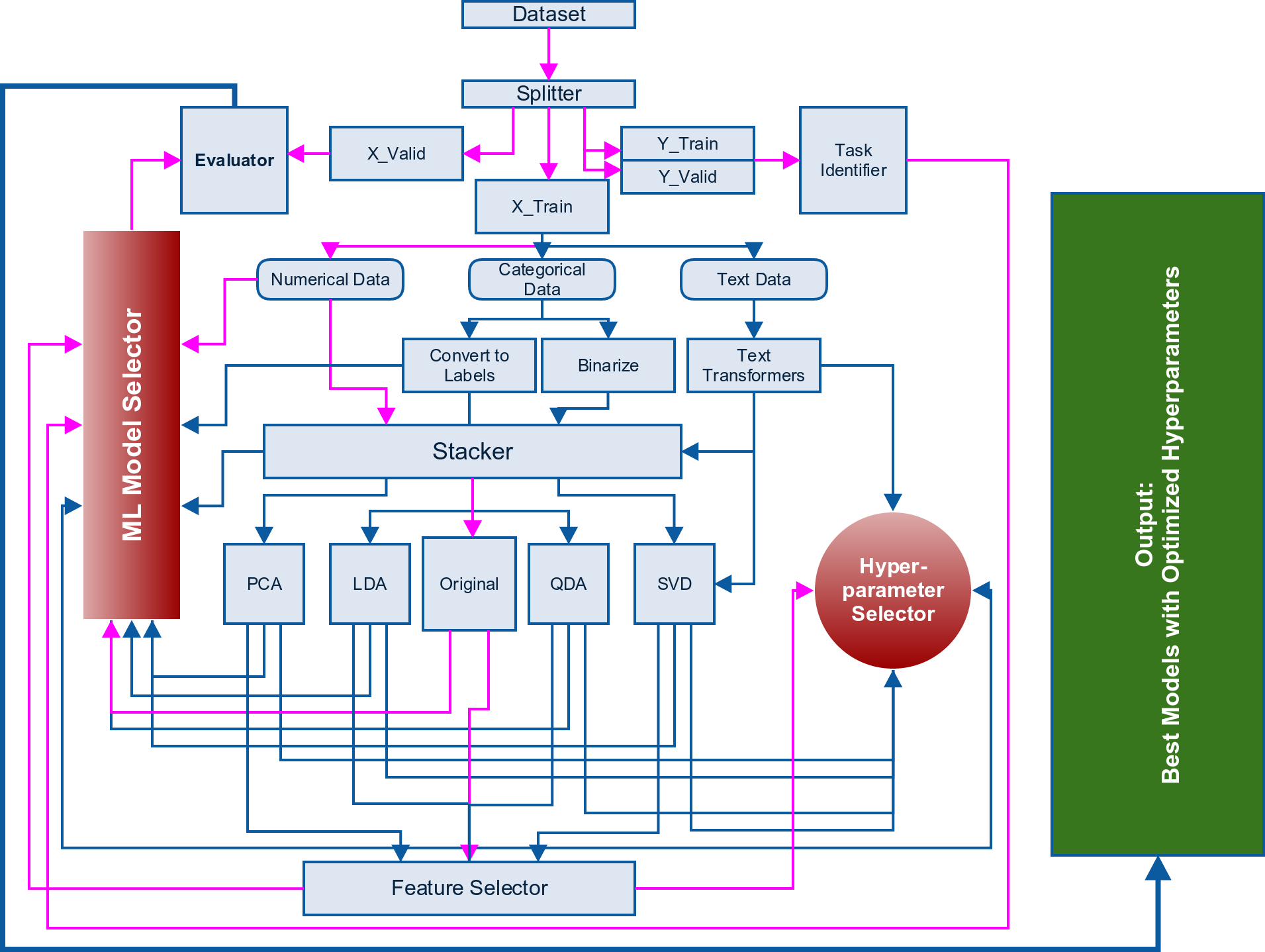}
\caption{Base framework of the proposed system. Pink lines represent the most common path followed.}
\end{figure}

The next section describes the most important components of the AutoCompete system in greater detail and also the strategy used for selection of models and tuning hyper-parameters.

\section{Components of AutoCompete}
\label{components}

The Dataset component of the AutoCompete framework receives the data from user in a tabular form. The data is splitted into training and validation set using the Splitter component. Various identifiers then identify the type of data and pass it to different pipelines and preprocessing steps. At every stage, the dataset is sent to ML Model Selector for model selection and evaluation. The Stacker takes the different types of preprocessed features and stacks them into one dataset for further decomposition and feature selection. Feature selection is also performed on the original dataset. The final output is the best pipeline with highest score or lowest loss in the evaluation.

Two major components of the proposed AutoCompete framework are the ML Model Selector and Hyper-parameter Selector, as highlighted in Figure 2. Table 1 shows the different classification and regression algorithms currently used by the AutoCompete framework. In addition to the modules specified in Table 1, we also introduce bagging and boosting for different models for improved performance at a later stage.

\begin{table}[H]
\caption{Classification and regression modules present in the current AutoCompete framework}
\begin{center}
    \begin{tabular}{ | l | l |}
    \hline
    \textbf{Classification} &  \textbf{Regression} \\ \hline
    Random Forest & Random Forest  \\ \hline
    Gradient Boosting &  Gradient Boosting \\ \hline
    Logistic Regression & Logistic Regression \\ \hline
    Ridge Classifier & Ridge  \\ \hline
    Naive Bayes & Lasso  \\ \hline
    SVM &  Support Vector Regression\\ \hline
    Nearest Neighbors & Linear Regressor  \\ \hline

    \end{tabular}
\end{center}
\end{table}

We propose two different selectors for selection of model and the corresponding hyper-parameters: (a) random search, (b) grid-search on a given parameter space. For both random search and grid search a parameter space is specified in the AutoCompete module according to different types of datasets encountered in the past. 

For example, in case of a text dataset, the modules selected are Term Frequency - Inverse Document Frequency followed by a decomposition method such as Singular Value Decomposition. After the decomposition process, the models Random Forest \citep{rf} and Support Vector Machines \citep{svm} are selected for initial results. To make the system fast, we tune only certain hyper-parameters and have a specified search space for these parameters. In case the Random Forest module is selected, we limit our search to number of estimators, minimum number of samples at each split and maximum number of features to be used by each estimator. Similarly, in case of SVMs, the kernel is fixed to radial basis function (rbf) and only the penalty parameter and gamma (kernel coefficient) is tuned. 

It is observed that even though we limit our system to tuning only certain parameters, we get results comparable to systems like hyperopt \citep{hyperopt} (these results have been discussed in the Experiments section) and also the results are obtained faster.

\section{Experiments}
\label{sec:exp}

We tested our framework on standard datasets such as MNIST \citep{mnist}, newsgroup-20 \citep{newsgroup}, adult dataset, smartphone dataset for human activity prediction \citep{smartphone} and housing dataset. These five datasets selected differ a lot from each other in terms of the number of variables, kind of data, machine learning task type to be applied and selection of evaluation metrics. They, thus, form a nice benchmark that can be used to develop other AutoML algorithms and frameworks on. Table 2 shows the different parameters for the datasets used.

\begin{table}[H]
\caption{Datasets used for testing AutoCompete framework}
\begin{center}
    \begin{tabular}{ | l | l | p{5cm} |}
    \hline
    \textbf{Dataset} &  \textbf{No. of Variables} & \textbf{Task Type} \\ \hline
    MNIST & 784 & Multiclass Classification \\ \hline
    Newsgroup-20 & \textasciitilde 100k & Multiclass Classification   \\ \hline
    Adult & 14 &  Binary Classification\\ \hline
    Smartphone & 561 & Binary Classification   \\ \hline
    Housing & 14  & Regression \\ \hline
    \end{tabular}
\end{center}
\end{table}

Results on adult dataset with a much smaller number of variables are presented first. For a small dataset like this one, AutoCompete selects a few fast models and then optimizes the hyper-parameters for the model with highest area under the ROC curve. AUC is chosen as the evaluation metric since the labels are skewed and a threshold on predicted probabilities will be more intuitive than classification accuracy. 

Figure 3 shows the models which were evaluated and their performance on the Adult dataset. The selected model with a grid based hyper-parameters for small dataset gives an AUC of 0.88.

\begin{figure}[H]
\centering
\includegraphics[scale = 0.3]{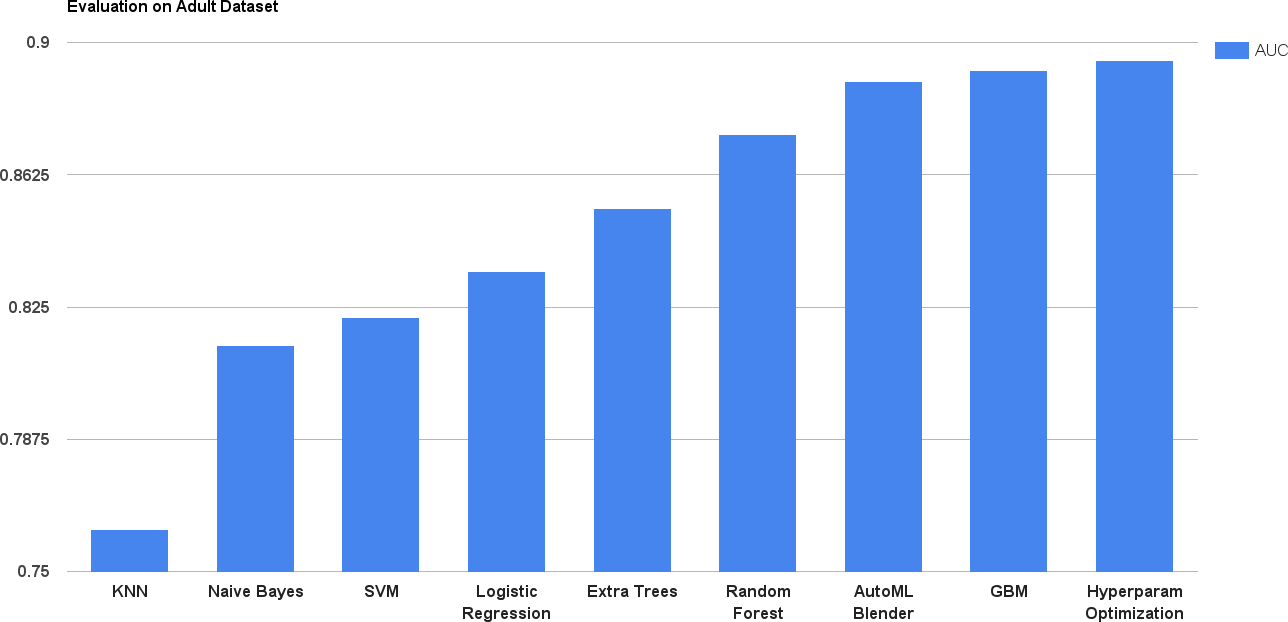}
\caption{ROC AUC for different model evaluations on the Adult dataset.}
\end{figure}

For MNIST, the parameters were chosen using both random search and grid search. A pipeline with PCA was selected with Random Forest as the model as prior information about the type of data is available to us. The accuracy on the test dataset was reported to be 0.96. Current framework is limited to 30 minutes of wall time and models are not evaluated further if this limit is reached. 

\begin{table}[H]
\caption{Results on MNIST dataset}
\begin{center}
    \begin{tabular}{ | l | l |}
    \hline
    \textbf{Algorithm} & \textbf{Accuracy Score}  \\ \hline
    Convnets & 99.8\%   \\ \hline
    hyperopt-sklearn & 98.7\%    \\ \hline
    libsvm grid-search & 98.6\%\\ \hline
    \textbf{AutoCompete} &    \textbf{96\%}\\ \hline
    \end{tabular}
\end{center}
\end{table}

In case of Newsgroups-20 dataset, the AutoCompete framework takes less than 10 minutes of wall time to beat hyperopt's results \citep{hyperopt}. The pipeline chosen in this case was a text transformer (TF-IDF) and logistic regression. 

\begin{table}[H]
\caption{Results on Newsgroups-20 dataset}
\begin{center}
    \begin{tabular}{ | l | l |}
    \hline
    \textbf{Algorithm} & \textbf{Weighted Average F1 Score}  \\ \hline
    \textbf{AutoCompete} &    \textbf{0.864}\\ \hline
    hyperopt-sklearn &  0.856   \\ \hline
    SVMTorch & 0.848 \\ \hline
    LibSVM & 0.843 \\ \hline
    
    \end{tabular}
\end{center}
\end{table}

The next two datasets, we tested AutoCompete framework on were the smartphone dataset and housing dataset. The smartphone dataset is a classification dataset and housing dataset on the other hand is a regression dataset. Smartphone dataset consists of 561 variables and all of them are numeric and housing dataset consists of 14 attributes which are a mixture of categorical, integers and real numbers. The selected pipeline and scores obtained on evaluation for all the datasets are shown in Table 5.

\begin{table}[H]
\caption{Selected pipeline and evaluation score for different datasets}
\begin{center}
    \begin{tabular}{ | l | l | l |}
    \hline
    \textbf{Dataset} & \textbf{Selected Pipeline} & \textbf{Evaluation Score} \\ \hline
    Smartphone & Logistic Regression & 0.921 (AUC)  \\ \hline
	Housing & RF(Features) + SVR & 2.3 (RMSE) \\ \hline 
	MNIST & PCA + RF & 0.96 (Accuracy) \\ \hline
	Newsgroup-20 & TFIDF + LR & 0.864 (Weighted F1) \\ \hline
	Adult & Model Stacker & 0.85 (AUC) \\ \hline   
    \end{tabular}
\end{center}
\end{table}

We also used AutoCompete in the AutoML Challenge. For the challenge, the AutoCompete system did not require any human interference. We ranked 2nd in the Phase0 of the competition. Since the AutoML phase required python code submission, which is still under development for AutoCompete, we did not participate in that phase. This will be incorporated and AutoCompete will be used in all the upcoming phases of the AutoML challenge. The results are shown in Figure 4.

\begin{figure}[H]
\centering
\includegraphics[scale = 0.35]{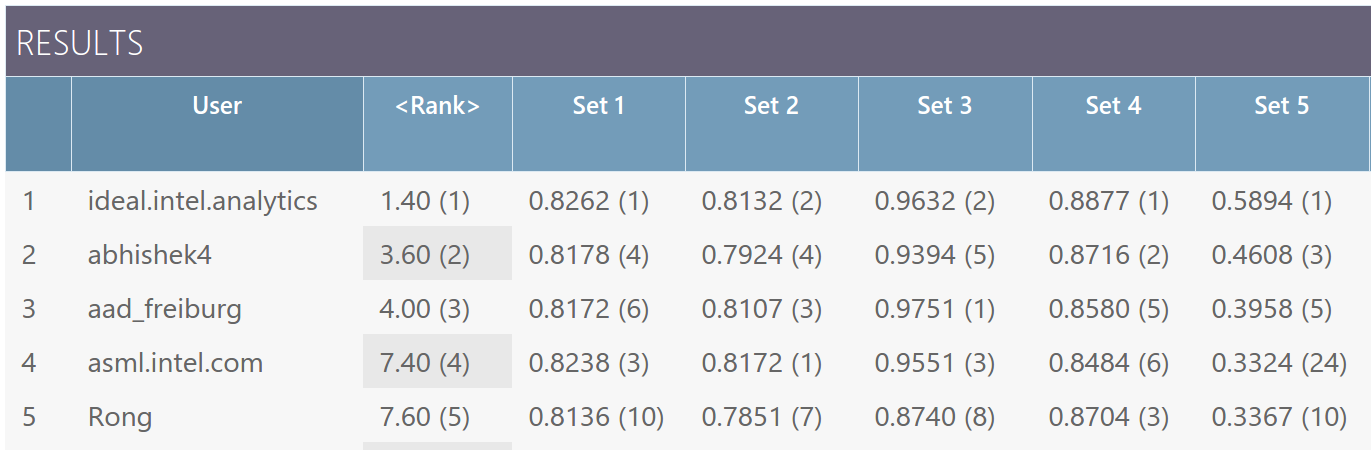}
\caption{Our result in the AutoML challenge.}
\end{figure}

All the computations were performed on a laptop with 4th gen Intel Core i7-4650U Processor (3.3 GHz, 4M Cache) and 16 GB RAM without any GPU power.

\section{Conclusion and Future Work}
\label{sec:conclu}

We introduce a highly automated framework for tackling machine learning problems. The framework and all pipelines inside were learned and designed based on the experience obtained by taking part in hundreds of machine learning competition over a period of two years. The comparison of AutoCompete with well established frameworks like hyperopt \citep{hyperopt} tells us that there is a high potential in AutoCompete in terms of minimizing human effort and bringing machine learning to masses. The proposed framework enables a novice in machine learning to create and build benchmarks for tabular datasets without much (or any) intervention. It is also seen that the system performs nicely on machine learning challenges. The underlying implementation is based purely on Python and scikit-learn \citep{sklearn} with some modules written in Cython. 

To extend the research in this field, our next steps (currently under research) would be to include a gender based genetic algorithm (GGA) \citep{gga}, Sequential Model-based Algorithm Configuration \citep{smac} and TPE \citep{tpe} for both selection of the machine learning model and tuning the hyper-parameters. Our future research also includes better stacking, ensembling of models and model blending to optimize for a required evaluation metric. We plan to release a usable version in the future and it will be available on the website of our research group. 

\bibliography{jmlr}
\end{document}